\pdfoutput=1
\documentclass[11pt]{article}

\usepackage{emnlp2021}

\usepackage{times}
\usepackage{latexsym}

\usepackage[T1]{fontenc}
\usepackage[utf8]{inputenc}

\usepackage{microtype}

\usepackage{dirtytalk}
\usepackage{amsmath,amssymb,mathtools}
\usepackage{booktabs}

\usepackage{graphicx}
\graphicspath{ {./graphics/} }

\DeclareMathOperator{\softmax}{softmax}

\title{A Temporal Variational Model for Story Generation}

\author{David Wilmot \\
  School of Informatics\\
  University of Edinburgh \\
  \texttt{david.wilmot@ed.ac.uk}\\
  \And
  Frank Keller \\
  School of Informatics\\
  University of Edinburgh \\
  \texttt{keller@inf.ed.ac.uk}\\
  }

\begin{document}

\maketitle
\begin{abstract}
Recent language models can generate interesting and grammatically correct text in story generation but often lack plot development and long-term coherence. This paper experiments with a latent vector planning approach based on a TD-VAE (Temporal Difference Variational Autoencoder), using the model for conditioning and reranking for text generation. The results demonstrate strong performance in automatic cloze and swapping evaluations. The human judgments show stories generated with TD-VAE reranking improve on a GPT-2 medium baseline and show comparable performance to a hierarchical LSTM reranking model. Conditioning on the latent vectors proves disappointing and deteriorates performance in human evaluation because it reduces the diversity of generation, and the models don't learn to progress the narrative. This highlights an important difference between technical task performance (e.g. cloze) and generating interesting stories.
\end{abstract}

\section{Introduction}

There has been huge recent success using large language models such as BERT \citep{devlin-etal-2019-bert} and the GPT family  \citep{Radford2018ImprovingLU,radford2019language,brown2020language} in a diverse range of language understanding and language generation tasks. In story generation, recent work has focused on using neural language generation either directly using language models, seq2seq models \citep{Roemmele2016WritingSW}, or part of a hierarchical model \citep{fan-etal-2018-hierarchical}. However, these models often generate stories that lack both diversity and long-range coherence \citep{See2019DoMP, Holtzman2018LearningTW}.

A story is a series of events \citep{abbott2008cambridge}. But a story is not just a series of events. It involves \citep{ryan2014possible} sudden switches in the plot, contrasts between the goals and results of characters’ actions, self-contradiction; repetition of narrative sequences; and elements of the narrated happenings that have multiple meanings.  The plot also requires events be significant and unpredictable  \citep{schmid2003narrativity}  while transgressing seemingly impenetrable barriers \citep{Lotman1977TheSO}, e.g. a hero on a dangerous quest or an unlikely marriage. Writing a good story requires tracking a complex set of entities and events.

Some more recent neural work has tried to address this challenge by using a multi-step planning approach:  A story skeleton is generated using semantic role labelling (SRL; \citealt{fan-etal-2019-strategies}), then a surface sentence is generated from each point on the plan. Numerous alternatives to SRL for story generation plans have been tried, including keyword extraction \citep{yao2019plan,goldfarb2019plan}, compression \citep{xu-etal-2018-skeleton}, event representation \citep{Martin2018EventRF,Ammanabrolu_Tien_Cheung_Luo_Ma_Martin_Riedl_2020}, and for non-neural models, graph-based generation \citep{10.5555/2891460.2891543}. \citet{goldfarb-tarrant-etal-2020-content} additionally use a suite of individually trained re-rankers on top of a multistage SRL planning pipeline. 

Latent state-space models \citep{koller2009probabilistic} abstract away the relationship from the observed input and output into a probabilistic latent state. In time series modelling, autoregressive models infer within the original observation space ($x$). In contrast, state-space models learn and infer temporal relations in the abstract state ($z$) and then reconstruct the observed state ($x$). This is in principle similar to the story planning skeletons that use representations such as SRL. The limitation to using keywords or SRL is that the model has to commit to a specific representation apriori. In this paper, we propose to use latent representations instead. The advantage of using a latent state for planning is that the model can learn the most valuable features to infer in the planning process. These representations can be richer than a predetermined simplification such as keywords or SRL.

\begin{figure*}[htbp]
  \centering
  \includegraphics[trim=0 0 0 0,clip,width=0.98\textwidth]{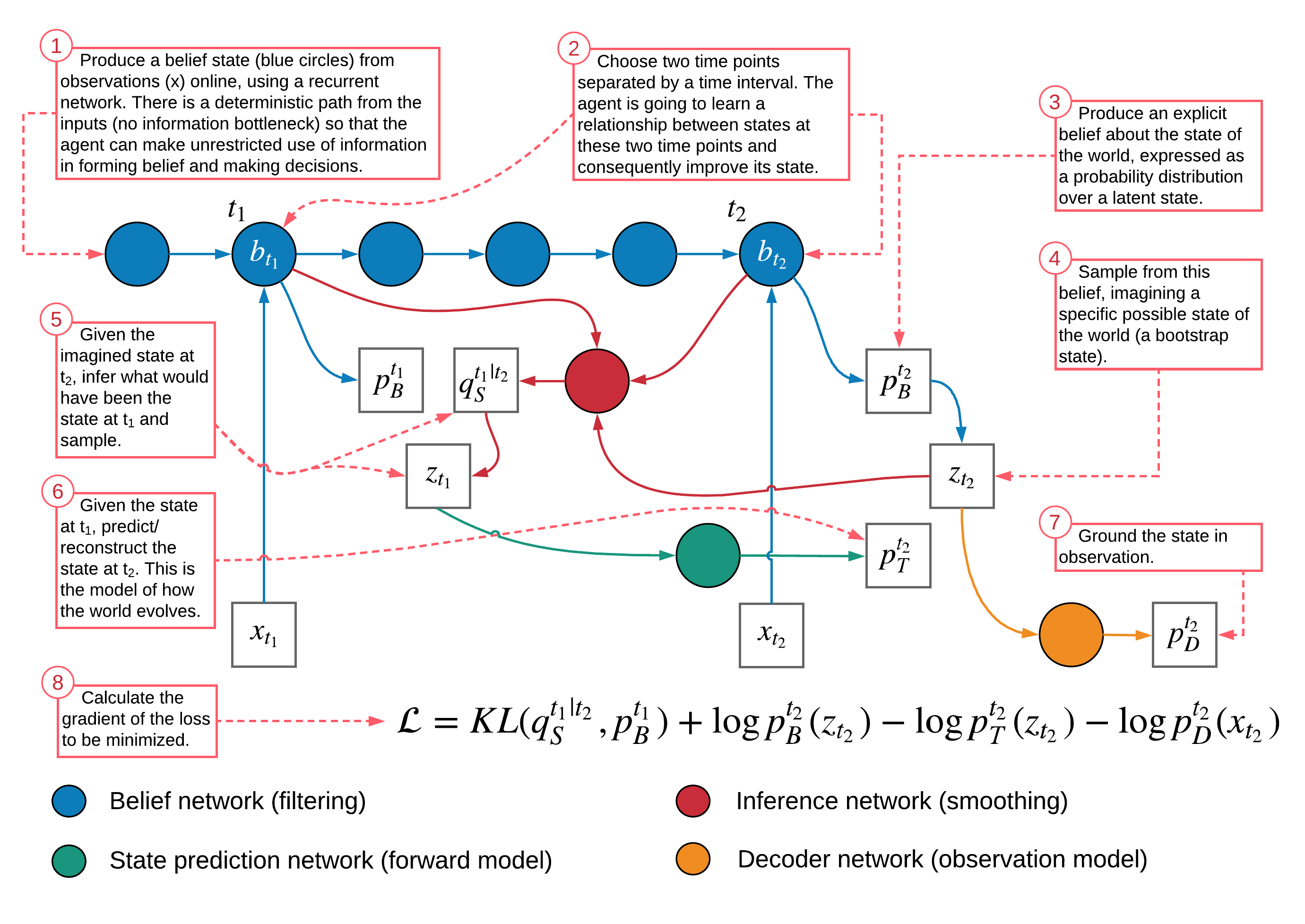}
  \caption{This is the main TD-VAE standalone architecture reproduced from \citet{gregor2018temporal}.}
  \label{fig:tdvae}
\end{figure*}

Specifically, we use the Temporal Difference Variational Autoencoder (TD-VAE; \citealt{gregor2018temporal}), a state-space model which learns an abstract state $z$, which is projected forward to a future state $z_{t_{n}}$ and then reconstructed to the input $x_{t_{n}}$. $t$ is a temporal representation referring to the index of a sentence within a story. For the internal workings of TD-VAE see fig \ref{fig:tdvae}. States space models have been the subject of much recent interest and development \citep{NIPS2013_d81f9c1b,NIPS2015_b618c321,NEURIPS2018_2de5d166,NEURIPS2019_b5d3ad89}. The intuition behind state-space models is they can learn an abstract state containing the most salient features of the environment, removing irrelevant details. These models have found much success in complex RL learning environments where the simplified latent state aids policy planning. While these models have been used for dialogue generation \citep{Serban2017AHL}, their use for other NLP tasks has been limited.

 In TD-VAE, $z$ is a Bayesian representation implemented using a variational autoencoder \citep{Kingma2014AutoEncodingVB}. Unlike autoregressive state-space models such as those proposed by \citet{DBLP:journals/corr/Graves13} or \citet{Doerr2018ProbabilisticRS} that only predict the next time step, TD-VAE can make jumpy predictions for multiple time steps ahead, which can reduce the errors from an autoregressive rollout. VAE models have been used in story generation and planning. \citet{Li2019LearningTW} use a conditional VAE that learns a latent global state cache which text generation conditions on. \citet{wang-etal-2020-plan} add a planning mechanism to a conditional VAE, but unlike their proposal, there is separate keyword skeleton story planning, and the planning isn't directly in the latent state.

Overall we apply our TD-VAE model to text generation in two ways: as a reranker and directly conditioning on the latent vectors for text generation. Our TD-VAE model when used directly as a discriminator to rerank generated candidate sentences \citep{Holtzman2018LearningTW} as part of a beam search of sampled continuations. To allow comparison with two-step planning methods, we also use Pseudo Self-Attention (PSA; \citealt{DBLP:journals/corr/abs-1908-06938}) to inject the expected latent state directly as history in the transformer model. This conditions directly on the latent vectors as a planning mechanism (\textit{TD-VAE Conditioning}), \citet{li-etal-2020-optimus} employs a similar approach. We find that the reranker model can improve text generation and perform well on automated coherence evaluations. The second approach, the conditioning model performs worse than other baselines because conditioning reduces the diversity of the text generated.

\section{Method}

\begin{figure*}[htbp]
  \centering
  \includegraphics[trim=0 70 130 0,clip,width=0.98\textwidth]{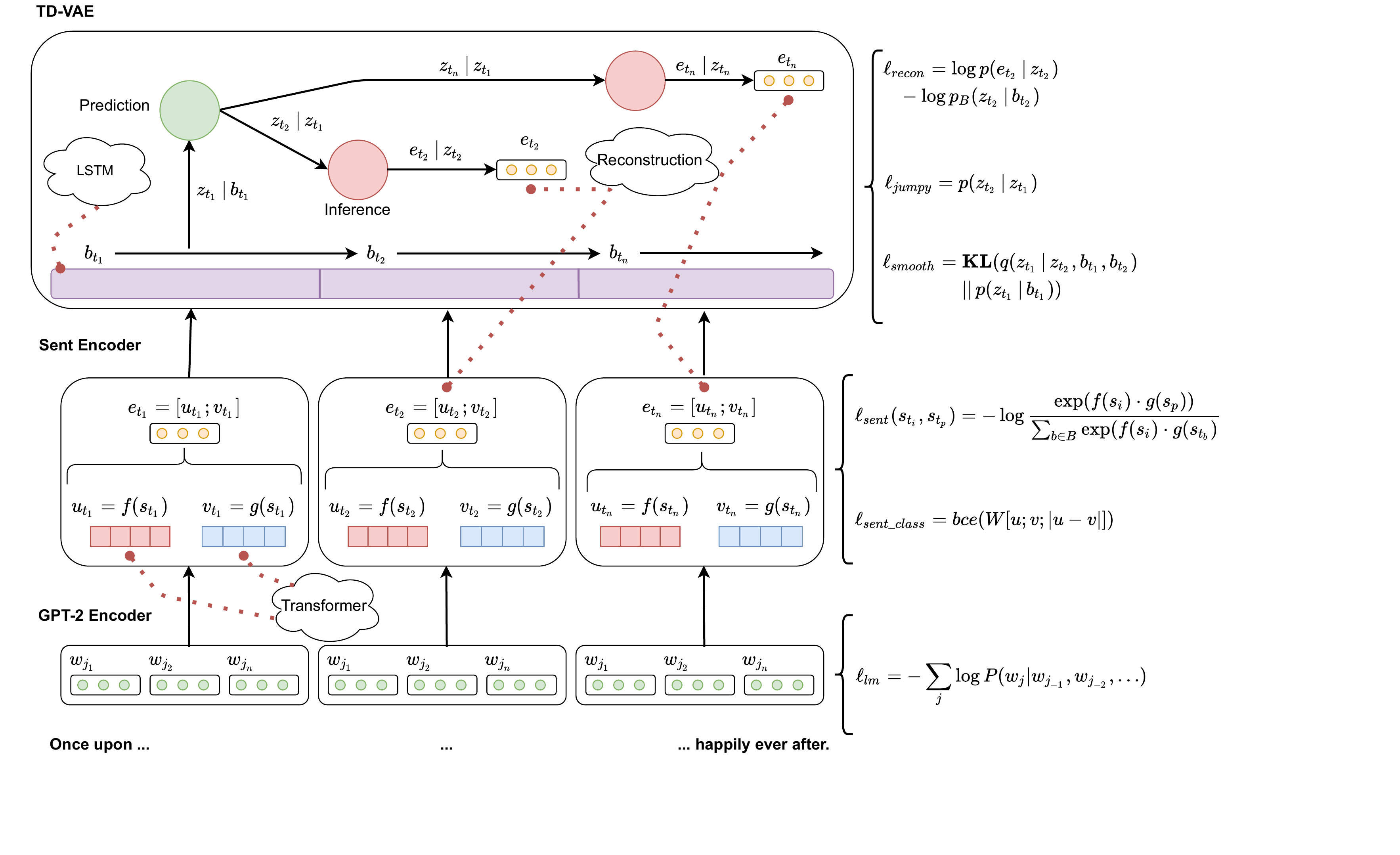}
  \caption{TD-VAE hierarchical architecture showing the multi-layer architecture: The base layer is GPT-2. From this, a sentence encoder infers sentence embeddings. The top-level is TD-VAE which learns to infer and reconstruct future sentence embedding states. It should be noted that that the sentence embeddings are concatenated when fed into the TD-VAE loss, but the respective loss maximises the dot product.}
  \label{fig:tdvae_hierarchy}
\end{figure*}

\subsection{Approach}

Our overall approach is to build a hierarchical model on top of GPT-2. On top of GPT-2 is a sentence encoder based on transformers that build rich sentence representations. The top level of the model is the TD-VAE, which learns to draw inference between the sentence time steps in the story. Figure~\ref{fig:tdvae_hierarchy} illustrates our architecture.\footnote{Code and configuration for this paper are available on Github at \url{https://github.com/dwlmt/knowledgeable-stories}.} 

\subsection{TD-VAE}

The top TD-VAE layer in the architecture, infers a future time step $t_2$ from an earlier time step $t_1$ (see Figure~\ref{fig:tdvae_hierarchy}). TD-VAE enables jumpy predictions of state, projecting forward multiple time steps of latent vectors from each position, conditioning from $t_1$ to $t_n$. The input is the sentence embedding $e_t$, which encapsulates the world's state as the reader observes it. These sentence representations are compressed into beliefs $b_{t_1}$ and $b_{t_2}$ using a stacked unidirectional LSTM. These beliefs are expectations about the state of the world in the future, i.e., they model what the reader expects to happen. These expectations aren't mapped directly to a future sentence state but via an abstract latent state $z_t$. The model can sample using variational inference from the latent distribution $p_B(z_{t_2}\,|\,b_{t_2})$ to guess the latent state for a future point in the story, and given the future latent state reconstruct the sentence embedding at this future point in the story, $p(e_{t_2}\,|\,z_{t_2})$. This allows the model to infer what the sentences at $2,3,4,\dots$ steps in the future will be like, without generating intermediate concrete sentence representations, and so functions as an abstract planning mechanism.

The reconstruction loss of the \textit{decoder} network is given in (\ref{eqn:tdvae_recon}). It is designed to maximise the probability of the sentence embedding reconstruction given the latent state at that time step. This is constrained by the second part, a bottleneck which prevents too much of the concrete sentence state $e_{t_2}$ being encoded in the latent state $z_{t_2}$.
\begin{equation}
\ell_{recon} = \log p(e_{t_2}\,|\,z_{t_2})  \\
- \log p_B(z_{t_2}\,|\,b_{t_2})
\label{eqn:tdvae_recon}
\end{equation}
To predict $t_2$, the model should estimate what the state of the world should be at $t_1$ for $t_2$ to have happened. To do that a \textit{smoothing} distribution is calculated $q(z_{t_1}|z_{t_2}, b_{t_1}, b_{t_2})$. The notion behind this is the forward prediction during training should be based on what did happen rather than unrealised possibilities. (\ref{eqn:tdvae_jumpy}) is the \textit{transition} distribution that projects the latent state forward into the future, it is the jumpy prediction:
\begin{equation}
\ell_{jumpy} = p(z_{t_2}\,|\,z_{t_1})
\label{eqn:tdvae_jumpy}
\end{equation}
Finally (\ref{eqn:tdvae_smoothing}) is the KL divergence between the smoothed state and what could have been known at $t_1$. Minimising this divergence indirectly trains the \textit{belief} distribution to learn what the state of the world should be at $t_1$ to anticipate it at $t_2$:
\begin{equation}
\ell_{smooth} = \mathbf{KL}(q(z_{t_1}\,|\,z_{t_2}, b_{t_1}, b_{t_2})\,||\,p(z_{t_1}\,|\,b_{t_1}))
\label{eqn:tdvae_smoothing}
\end{equation}
Combining these losses produces the overall loss:
\begin{equation}
\begin{split}
\ell({t_1, t_2}) = \mathbb{E}[
  \log p_D(e_{t_2}|z_{t_2}) \\
  + \log p_B(z_{t_1}|b_{t_1}) \\
  + \log p_T(z_{t_2}|z_{t_1}) \\
  - \log p_B(z_{t_2}|b_{t_2}) \\
- \log p_S(z_{t_1}|z_{t_2}, b_{t_1}, b_{t_2})]
\end{split}
\label{eqn:tdvae}
\end{equation}
Training is performed by taking $n$ samples randomly selected (default $100$) for each batch using a linear sample of time steps between $t_{+1}$ to  $t_{+k}$. This allows the model to learn to make jumpy predictions more than $t + 1$ steps ahead. We use the hierarchical version of the TD-VAE model where layers are stacked so that each layer samples from those below it.

\subsection{Sentence Encoding}

In the middle layer of the architecture (see Figure~\ref{fig:tdvae_hierarchy}), we encode a sentence representation for each sentence in the batch. The sentence representation needs to be independent and detached (from gradient propagation) in the TD-VAE layer above; otherwise, TD-VAE will force the sentence representations to be degenerate. 

The sentence representations are based primarily on Quick-Thoughts \citep{logeswaran2018an}, inspired by Skip-Thoughts \citep{NIPS2015_5950}. For the sentence representations, let there be two encoded vectors from two functions $u = f(s)$ and $v = g(s)$, with $s$ being the sequence of the word tokens representing the sentence output by GPT-2. $f(s)$ and $g(s)$ can be any functions that can convert a sequence of inputs into a single vector. Unlike Quick-Thoughts, $f(s)$ and $g(s)$ are stacked autoregressive transformers \citep{NIPS2017_3f5ee243} that use embeddings to encode position. The weights are not shared between the two encoders. To produce a single vector per sentence, $u$ and $v$ concatenated across the embeddings from the GPT-2 layer.\footnote{Mean pooling was all experimented with but proved inferior and is not used in any reported results.}

For a batch of sentences, the loss is for a given candidate sentence $s^{cand}$ is the negative log probability of the dot product between $f(s_{t_{2}})$ and $g(s_{t_{2}})$, normalised by all the other sentences in the batch, indexed by $i$, see~(\ref{eqn:quickthoughts_loss}). For each sentence in the batch there are two correct candidates $s_{t_{-1}}$ and $s_{t_{1}}$, given by the target position $s_{t_{p}}$ in~(\ref{eqn:quickthoughts_loss}), the previous and following sentence. The intuition behind this loss is it encourages vectors to be similar to their neighbours while dissimilar to sentences further away in the batch. Making representations in narratives more localised is desirable since it is specific events described by a sentence that we want the TD-VAE model to learn to project forward to plan the plot.  A custom sentence encoder is preferred to a model such as Sentence BERT \citep{reimers-gurevych-2019-sentence}, as it allows us to take advantage of the finetuning of the underlying language model.
\begin{equation}
\begin{aligned}
\ell_{sent}(s_{t_i},s_{t_p}
) = -\log \frac{\exp(f(s_{i}) \cdot g(s_{p}))}{\sum_{b\in B} \exp(f(s_i) \cdot g(s_{t_{b}})}
\label{eqn:quickthoughts_loss}
\end{aligned}
\end{equation}
To enrich the sentence representations, we finetune on SNLI datasets \citep{bowman-etal-2015-large,N18-1101} and the  common sense reasoning dataset ATOMIC \citep{Sap2019ATOMICAA}.  As per InferSent \citep{conneau-etal-2017-supervised}, we concatenate the sentence embedding vectors together and apply a single projection layer $W$, $\softmax(W[u;v;|u - v|])$, and train using binary cross-entropy. The complete sentence loss is $\ell_{sent} + \ell_{snli} + \ell_{atomic}$. The motivation for this finetuning is that supervised tasks have been shown to improve sentence representations \citep{DBLP:journals/corr/abs-1803-11175,reimers-gurevych-2019-sentence}, and both entailment and common sense inference are relevant to the story domain.

\subsection{Language Model}

The language model is Generative Pre-Training~2 (GPT-2; \citealt{radford2019language}), using the \textit{medium} model with $345$ million parameters as a base. For text generation, an auto-regressive language model is preferred to a bidirectional and masked one, such as BERT \citep{devlin-etal-2019-bert}. The model is fine-tuned using the original LM loss in~(\ref{eqn:lm_loss}) on the story datasets. $w_j$ is a particular encoded word piece, $w$, in a passage of text.
\begin{equation}
%\begin{aligned}
\ell_{lm} = - \sum_{j} \log P({w_j | w_{j_{-1}}, w_{j_{-2}}, \dots)}
\label{eqn:lm_loss}
%\end{aligned}
\end{equation}
Training GPT-2 works best on longer blocks of text whereas the hierarchical model, outlined later, encodes sentences as vectors and trains predictions based on the whole story. To train GPT-2 efficiently, the language model training is run distinct from the hierarchical training, with the training process alternating per batch between the two. This allows GPT-2 to be fine-tuned with a longer context of $1024$ word tokens.

\subsection{Conditional Generation}

$e_{t_{-n}}$ vectors for future states are calculated by sampling forward with the TD-VAE model to via the state space $z_t$ of the model and reconstructing a vector in the sentence embeddings space $e_{t}$.  To condition text generation on predicted sentence vectors, the $e_t$ sentence representations are projected into the GPT-2 transformer space. A hidden space $h_{mem}$ is created by applying a weight matrix $W_{m}e_t$, with $W_m \in \mathbb{R}^{L \cdot H \cdot P}$, where $P$ is the dimensionality of the sentence vector. TD-VAE projections $e_t$, $L$ is the number of layers in the GPT-2 transformers and $H$ the size of the transformer hidden state. The hidden state is shared across transformer heads; this method is Pseudo Self-Attention \citep{DBLP:journals/corr/abs-1908-06938}. The approach allows the hidden state to be prefixed directly onto the GPT-2 hidden state when generating or spliced into it when generating longer sequences. Thus the latent state can be used to control text generation. With models that use conditional generation, there is an additional training step. For alternating training iterations, the $e_t$ sentence representations are projected into the GPT vector space as described, concatenated before the GPT representation, and then fine-tuned using the LM objective. Thus the training process learns to use the latent vectors to improve text generation directly.

\subsection{Datasets}

WritingPrompts \citep{fan-etal-2018-hierarchical} is a dataset that consists of circa 300k short stories of circa 50 sentences in length from a Reddit creative writing group. Only WritingPrompts is used for evaluation in the following experiments, whereas the following datasets are used during training: CMU Books \citep{DBLP:journals/corr/abs-1305-1319} and Movies Corpora \citep{bamman-etal-2013-learning}, which are summaries for whole novels and movies, and the CBT dataset \citep{journals/corr/HillBCW15} of children's books. Longer story forms may also provide improvements over short stories, so the full processed text from Schmoop \citep{DBLP:journals/corr/abs-1912-13082}, the Books Corpus \citep{Zhu2015AligningBA} and Film Corpus \citep{Lin2011AllTW} are also used. In practice, models trained on multiple datasets performed better on automatic evaluation than single dataset trained models, and so only these results are presented in four experiments. For training, we use the existing dataset split for WritingPrompts.

We train models by sampling batches from each dataset in proportion to the dataset size. During training, batches of four blocks of $100$ sentences for each story were used. For the TD-VAE training per block, the model randomly samples $200$ example states to reconstruct where the target state is randomly sampled between $t+1$ and $t+5$ sentence ahead using a linear sampling scheme. SGD with Nesterov momentum was used to train the model with a learning rate of $0.01$ and momentum of $0.9$. Models were trained with 10,000 batches per epoch (note this is not the whole dataset), early stopping once the model failed to improve every three epochs, and the learning rate is halved every epoch without improvement.

\subsection{Implementation Details}

The models are implemented on the AllenNLP framework \citep{gardner-etal-2018-allennlp} on Pytorch \citep{NEURIPS2019_9015}. The final evaluated models used the finetuned GPT-2 medium model for the LM layer, six layers of transformer with a $1024$ embedding size (matching GPT-2) for the sentence encoding layer layers top layers of the respective models: LSTM/Transformer/TD-VAE. The primary TD-VAE model has $485$M tunable parameters, circa $345$M are from GPT-2.

\subsection{Planning Generation}

\begin{figure*}[htbp]
  \centering
  \includegraphics[trim=0 0 180 0,clip,width=0.98\textwidth]{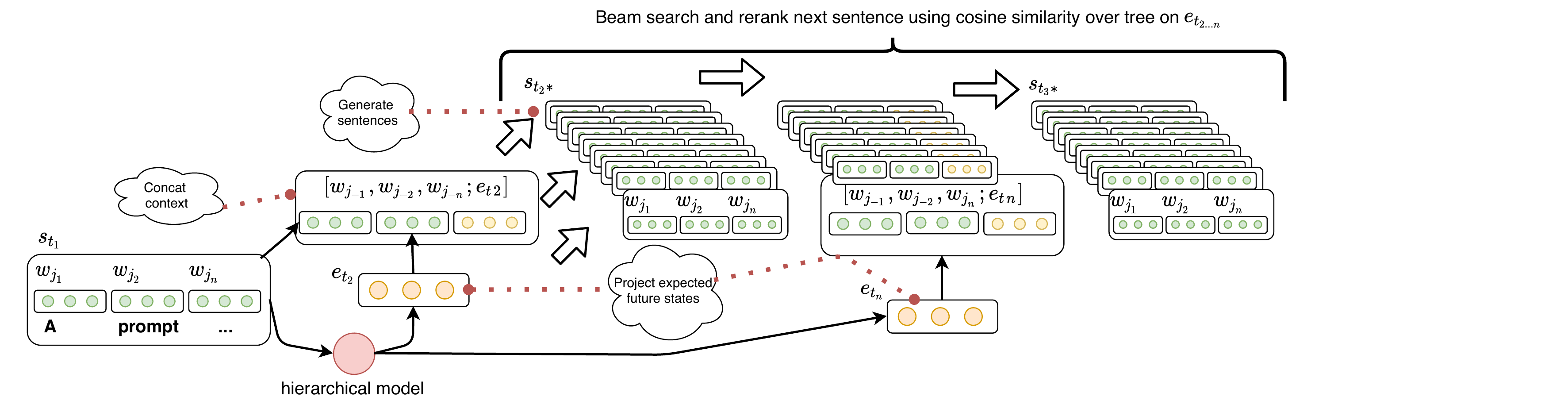}
  \caption{For conditional text generation, the expected $e_{t_2}$ vector is the TD-VAE inferred state for the following sentence. It is concatenated with the existing prompt, and sentences are sampled from the conditioned vector. This happens iteratively as part of a beam search (beam $10$) where the most likely sentences are kept based on how close they are to the expected $e_{t_2}$ vectors using cosine similarity. The final single story is the most like path through the beam search.}
  \label{fig:tdvae_writing}
\end{figure*}

Baseline text generation uses top-p sampling (or nucleus sampling) from \citet{Holtzman2020The} on the underlying GPT-2 model. This approach restricts the sampled tokens in generation to the top cumulative probability distribution tokens. Holtzman et al. found top-p sampling generates sentences with better coherence and diversity than top-k sampling \citep{fan-etal-2018-hierarchical,Holtzman2018LearningTW,radford2019language} methods. The default top-p threshold in our experiments is $0.925$. 

Most of the discrete plan-based methods generate text in two phases; the first generates a discrete plan for the stories, the second realises the generated text from the plan. Figure~\ref{fig:tdvae_writing} illustrates how this can be extended with TD-VAE and latent planning. Given a prompt, TD-VAE can project forward the most likely sentence vectors to $n$ steps in the future. These vectors can then be conditioned on using the projection layer described by splicing the projected vector into the hidden state history of the transformer. The advantage is that the latent state information is incorporated as a natural part of the GPT-2 history, and so architectural changes are not required for GPT. Most planning approaches generate a template for the whole story in advance, but this implementation is incremental: TD-VAE projects forward sentences over $n$ sentences incrementally, choosing the next sentence and then projecting forward again. 

In addition to conditioning on these latent vectors, we employ a beam search of $k$ (10 in evaluated stories) reranking approach that generates $n$ (100 per time step for the evaluated stories) candidates sentences and then reranks them by minimising the reconstruction error cosine similarity with the reconstruction from the latent state $z$. For text generation, both conditioning and reranking can be applied independently. However, in practice, conditioning worked better when combined with sampling multiple candidate sentences and reranking.  Only stories generated with either reranking or reranking with conditioning are judged in the human evaluation.

\subsection{Baselines}

The first baseline is the GPT-2 model without the hierarchical layers above it. Additionally, two other architectures have been used with the same hierarchical setup but using an LSTM \citep{HochSchm97} or Transformer \citep{NIPS2017_3f5ee243} discriminator instead of TD-VAE at the story level. These models are trained to discriminate between the following correct sentence and all the other sentences in the batch by maximising the correct dot product versus the incorrect one between the sentence representation and the encoded vector state of the LSTM or the transformer. The loss is the same as the $\ell_{disc}$ loss, except that it is the concatenated $e_t$ vector used with the LSTM or Transformer sentence vector as $u$ and $v$. This is the same loss as used by \citet{wilmot-keller-2020-modelling} for modelling story suspense. Story generation is performed by sampling $n$ sentence using GPT and reranking for the following sentence using the softmax over the vector dot products of the candidates as a probability distribution.

\section{Experiments}

\subsection{Automatic Structural Evaluation}

\begin{table*}[t]
\centering
\begin{tabular}{ccccccc}
\toprule
\textbf{Task} & \textbf{Model} &  \textbf{Easy} $\mathbf{\pm}$ \textbf{(CI)} & \textbf{K-1} $\mathbf{\pm}$ \textbf{(CI)} & \textbf{K-5} $\mathbf{\pm}$ \textbf{(CI)} & \textbf{K-10} $\mathbf{\pm}$ \textbf{(CI)} & \textbf{Avg} $\mathbf{\pm}$ \textbf{(CI)} 
 \\ \midrule
\textbf{Swap 2}  & Rand                        &   .500 (.050)   & .037 (.020)        & .190 (.040)        & .377 (.049)  & .276 (.045)
\\

- & LM                                 & .545 (.050)  & .045 (.022)         & .239 (.043)        & .473 (.050) & .326 (.047) \\
- & LSTM                                & .946 (.024)  & .035 (.019)        & .226 (.042)        & .441 (.050)  & .412 (.049)        \\

- & TRANS                                  & .953 (.022)   & \textbf{.058 (.024)}         & .238 (.041)        & .427 (.049) & .419 (.049)  \\
- & TD-VAE                                 & \textbf{.979 (.000)}   & \textbf{.058 (.024)}         & \textbf{.279 (.045)}         & \textbf{.496 (.050)}         
&  \textbf{.453 (.050)} \\
\midrule

\textbf{Mut 1} & Rand                                 & .500 (.050)  & .018 (.014)         & .094 (.030)         & .188 (.039) & .200 (.040) \\

- & LM                                 & .640 (.048)   & .026 (.017)        & .121 (.033)        & .255 (.044)  & .261    (.044)     
 \\

- & LSTM                     & .932 (.026)   & .106 (.031)       & .187 (.039)        & .303 (.046)  & .382 (.048)
 \\ 

- & TRANS                                 & \textbf{.966 (.019)}   & .020 (.015)        & .217 (.042)        & .379 (.049) & .396 (.049)    
 \\

- & TD-VAE                                 & .931 (.026)   & \textbf{.134 (.035)}         & \textbf{.356 (.048)}         & \textbf{.556 (.050)}  & \textbf{.494 (.050)}       \\

 \bottomrule
\end{tabular}
\caption{Results of the hard cloze and swap tasks for models trained on all datasets. Confidence Interval at 0.05. Averages are means across all the other tasks.}
\label{tab:cloze_res_hard_all}
\end{table*}

BLEU \citep{xu-etal-2018-skeleton}, and similar metrics are problematic for evaluating story generation, as is it unlikely there will be much overlap between well-written stories, even if they share a common prompt. ROC cloze endings have been widely used \citep{schwartz-etal-2017-effect} for evaluation, and we also use cloze and swapping for automated structural evaluations.  For swap, we randomly swap two sentences in each story. For mutation, we randomly sample a sentence generated by GPT-2 at that point in the story, conditioned on the story up to that point; the rationale is that just randomly selecting a cloze alternative sentence makes it too easy. There are two versions of the task: In the easy version, the task is to distinguish the original from the modified version of the story.  For the harder version, the task is to identify which of the sentences have been modified (mutated or swapped) by determining if they are in the top $K$ least likely candidates across the whole story with varying $K$ values reported.  

The results are given in Table~\ref{tab:cloze_res_hard_all}. Random results are obtained by randomly choosing the $K$ most likely sentences. The evaluation is on 400 random stories of length between 25 and 75 sentences from the WritingPrompts testset. The LM models results are calculated using the perplexity of a two-sentence sliding window. Each of the other models' probabilities are based on the softmax of a distance metric. Therefore, the mutated or swapped sentence should be the furthest away from expectations using cosine distance. For the hierarchical LSTM, the transformer discriminator is used to predict the correct answer as per conditional generation, and for the LM, lower perplexity is used.

The easy story cloze task has strong results of greater than 0.9 for most LSTM, transformer and TD-VAE models in telling the original from the modified story, so this task is not that useful. The language model (LM) performs much worse in identifying the original story, so all the hierarchical models are improved.  This is perhaps not surprising with cloze, as sampling from the same LM makes it unlikely this will perform well in detecting its own sampled sentences from the original. However, the comparison with the hierarchical models is tougher: For the hard task, K-1, K-5, and K-10, the TD-VAE model shows a clear improvement in both the transformer and LSTM models the swap and cloze tasks. This shows that TD-VAE is able to infer coherence well and is an improvement on the baseline hierarchical models.

\subsection{Generated Story Evaluation}

\begin{figure*}[t]
  \centering
  \includegraphics[width=0.98\textwidth]{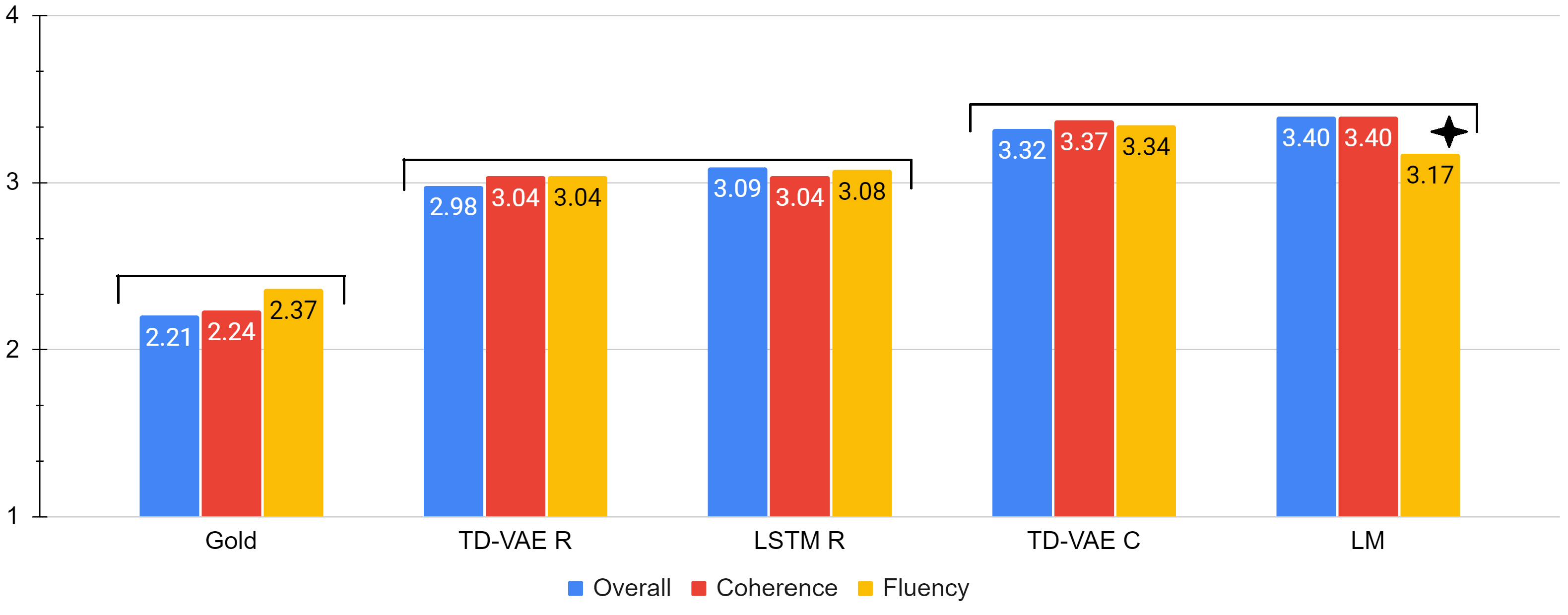}
  \caption{Continuation story generation results. Bar markers split into groups at $p < 0.05$ significance using a pairwise t-test. The exception is LM fluency marked with the star, which is statistically the same as LSTM R and TD-VAE R.}
  \label{fig:cont_bar}
\end{figure*}

Inferring coherence is quite different from the model's influence on story generation, which we evaluate using human judgements on MTurk. We take the writing prompt and the first $20$ sentences for $50$ stories. For each of these stories, we generate continuations of $5$ sentences. We evaluate using these long contexts because some initial tests of generating long stories from just the writing prompt found that the stories generated by the different models varied so much that it was hard for annotators to compare them reliably. A shorter extension to a longer context also requires the model to have a strong understanding of the long context to generate a coherent continuation. 

Five models are evaluated: \textit{Gold}, the actual human-written story from the dataset: \textit{LM}, the fine-tuned GPT-2 medium model, \textit{LSTM Reranking}, which generates multiple sentences using GPT and reranks and selects the most likely candidate using a beam search. LSTM Reranking is used instead of the transformer model because the automated benchmarks are similar, so evaluating both would be superfluous. \textit{TD-VAE Reranking} used beam search like the LSTM Reranking model but based on the TD-VAE. \textit{TD-VAE Conditioning} differs in that it directly conditions on the expected latent vectors. So the TD-VAE expectations change the text generated, unlike the Reranking model, which only filters out less suitable sentences generated by GPT-2. Mechanical Turk workers are asked to rank the five options from best to worst using three criteria: \textit{Overall} (a subjective judgement on their overall impression of the story.) \textit{coherence} (internal consistency of characters, places, actions and plot development.) \textit{Fluency} (how well written the story is, including its use of expressive language.) 

Additionally, MTurk workers were asked to summarise the story and give reasons for their overall choice. This served to filter poor quality submissions and to give insight into the factors they considered important. Five annotators judged each story after screening out and re-collecting for low-quality annotations.  Overall inter-annotator agreement for the task is $0.35$ using Krippendorff's alpha; this is fairly low, but the task is inherently subjective. 

Figure~\ref{fig:cont_bar} shows how the models were ranked by annotators on a scale of 1 to~5, i.e., lower is better. There isn't much difference between the different questions (overall, coherence, and fluency), except that fluency is relatively better for the base GPT-2 language model than overall and coherence. The gold standard stories are better than all model-generated continuations. The two reranking models LSTM and TD-VAE R, are significantly better than the base LM model, which means reranking improves the story over straight sampling. The TD-VAE C model that uses the planning mechanism doesn't improve the base LM, and fluency is worse. 

\begin{table*}[t]
\centering
\begin{tabular}{@{}lcccccc@{}}
\toprule
\textbf{Model} & \multicolumn{1}{l}{\textbf{Nouns}} & \multicolumn{1}{l}{\textbf{Verbs}} & \multicolumn{1}{l}{\textbf{Coreferences}} & \multicolumn{1}{l}{\textbf{Coreference Chains}} &
\multicolumn{1}{l}{\textbf{Meteor}} &
\multicolumn{1}{l}{\textbf{BLEU}}\\ \midrule
Gold           & \textbf{4.08}                      & \textbf{2.04}                      & \textbf{2.23}                      & 4.17       & - & -            \\
LM             & 2.48                               & 1.29                               & 2.06                               & 4.69          & .101                                & .628                                      \\
LSTM           & 2.59                               & 1.21                               & 2.18                               & \textbf{5.02}    & \textbf{.107}                                & .619                           \\
TD-VAE R       & 3.19                               & 1.41                               & 1.61                               & 4.00       & .090                                & \textbf{.705}                             \\
TD-VAE C       & 2.07                               & .885                               & 1.65                               & 3.99   & .083                                & .639                              \\ \bottomrule
\end{tabular}
\caption{Statistics with the average number of unique nouns and verbs (after stemming), coreferences per 100 tokens, and length of coreference mention chains per story. Meteor and BLEU are measured against the Gold stories. These stats are on a more extensive set of 400 stories and text of up to 2000 characters per generation.}
\label{tab:continuation_story_diversity}
\end{table*}

The main question is why TD-VAE Conditioning can perform well on automated evaluation (close and swap) but not improve story generation compared to the LSTM-based reranking model. Inspection of the stories used for evaluation suggests that the conditioning model is more repetitive than other models and uses more repeated linguistic constructs in continuations. We analysed the linguistic diversity of the generated continuations by counting unique nouns and verbs (see Table~\ref{tab:continuation_story_diversity}). We found that the TD-VAE C model has far less diversity than any other models and differs markedly from the human-written stories. For both TD-VAE models, the number of unique coreferences is lower than other models. In contrast, the coreference chain length is similar, implying that the TD-VAE models are more repetitive when generating coreferences and are less likely to introduce new entities, again reducing text diversity. On the other hand, TD-VAE R outperforms both the LM and LSTM model in terms of noun and verb diversity. The reduction can be seen in the appendix examples in \ref{sec:example1}.

For completeness, we also report Meteor and BLEU scores (also in Table~\ref{tab:continuation_story_diversity}); we find the TD-VAE models outperform the LM baseline and the LSTM-based hierarchical model in terms of BLEU, while the patterns is reversed for Meteor. We already pointed out the problem with using these metrics for story evaluation.

If a planning mechanism works, we would expect it to increase lexical diversity, as the planning would condition the LM to explore news areas and develop the plot. In contrast, it would appear the opposite is happening here. It has been noted by \citet{Welleck2020Neural} amongst others that large language models can produce dull and repetitive text and often copy from the context. The failure of the conditioning model could be an analogue, where TD-VAE can train well by expecting the future text to be similar in topic and style to the text seen. This would lead conditioning in the latent vectors to produce more dull and repetitive text than the LM on its own, which appears to be the case for TD-VAE~C. There seems to be a clear split between TD-VAE used in a ranking model and when conditioned on. The strong automated performance and improvement on the base LM with human evaluation show that TD-VAE~R can determine coherent continuations and improve story generation. However, when conditioned on this, it produces text more similar to the existing text, not interesting and eventful plots. Part of this could come from the richness of the latent representations: In a keyword planning system, the plot skeleton conditioned on is quite loose, for example from \citet{wang-etal-2020-plan}: \textit{lake} $\rightarrow$ \textit{friends} $\rightarrow$ \textit{water} $\rightarrow$ \textit{swim} $\rightarrow$ \textit{shore}. This allows many possible realisations and is not that restrictive of the LM. In contrast, sentence embeddings represent many more linguistic and semantic properties, so learning to condition them can cause generated text to be repetitive and limited.

\section{Conclusion}

Overall this paper has shown that the TD-VAE model can outperform alternative LSTM and transformer models on automatic coherence evaluation. When used for reranking, we also find that the TD-VAE model can improve text generation of a baseline LM. However, the conditioning TD-VAE model does not improve in a baseline because the latent vector conditioning reduces the diversity of text generated, which is crucial for exciting storytelling. While the ranking model shows some improvement and the models show it can improve comprehension, latent planning did not work as intended to produce better stories. If it had, then further comparison with SOA multi-stage planning or newer VAE text generation models would have been warranted with human evaluation. A warning to emphasise in this paper shows that improvements in technical measures such as BLEU or cloze can worsen generated text. \citet{guan-huang-2020-union} and \citet{guan-etal-2021-openmeva} have found many of these automated measures, including ROUGE, Perplexity for stories, and more sophisticated ones such as BERT-Score, correlate poorly with human judgement. 

In narrative text, there are continual shifts in the text in terms of topic, events, characters, and places that are hard to anticipate from sentence to sentence. Simply inferring coherence is not enough; instead, the generation process must push the story in new directions while remaining coherent to what has gone before. The challenge for future work is to adapt latent state models to focus on the plot progression rather than keeping the benefits of having a model self-learn the most salient representations. One avenue for this may the success of recent discrete VQ-VAE models \citep{Razavi2019GeneratingDH,NIPS2017_7a98af17}. Just as SRL or keyword-based planning forces a simplified representation, VQ-VAE or similar discrete models may force a simplified representation. This could make learning plot progression more straightforward and hence improve the quality of generated stories.

% The paper has shown some positive results for a novel model in modelling coherence, and when used to enhance an existing LM via reranking, it can improve generation. Yet, the challenge of story plotting is much harder, and while latent state models have promise, the models will have to be rethought to encourage better modelling of plot and progression.

\section*{Acknowledgments}

We like to thank the Mechanical Turk workers who evaluated the generated stories. Wilmot's work is funded by an EPSRC doctoral training
award.

\section*{Ethics Statement}

The evaluation study reported in Section 3.2 involved
participants recruited through Amazon Mechanical Turk. All
participants gave informed consent, received appropriate
compensation, and were informed that participation was
voluntary. Some texts could contain mildly offensive language,
which participants were also informed about before taking part.
The study had received prior approval by the ethics panel of the
School of Informatics at the University of Edinburgh.

\bibliography{main}

\clearpage

\appendix

\label{sec:appendix}

\section{Story Continuation Evaluation}

\subsection{Instructions}

\textit{The following are the instructions provided to the MTurk Annotators}:

Read a writing prompt and the first part of a story. Then read 5 different continuations of the story. Rank the story continuations from 1 (Best) to 5 (Worst) for a set of criteria. Judgements will be subjective but try to judge each according to the criteria given. You are only ranking how well each continuation follows on from the story, not the story itself. You may do multiple HITS with different stories but please do no more than 20 in total. Estimated time is 10 minutes per HIT. The prompts often begin "[WP]" which is just a convention used on the forum the stories are collected from. Rank the stories according to the following criteria:

\begin{itemize}
  \item \textbf{Coherence}: Narrative coherence is the degree to which the continuation makes sense given what has gone before. Coherent stories are internally consistent, and involve a plot development and progression with characters, places and actions that are meaningful in the context of the prior story.
  \item \textbf{Fluency}: How well does the continuation follow on from the story, is it natural, does it follow in terms of style, grammar, use of literary devices, and expressive language.
  \item \textbf{Overall}: This is an overall judgement of how good and interesting you think the continuations are in terms of criteria such as character, plot and how well written they are.
\end{itemize}

Note, the rankings for each continuation must be unique for each question category - Overall, Coherence, and Fluency. No two continuations can be given the same rank, and so one continuation must get each of the ranks 1-5.

\begin{figure*}[htbp]
  \centering
  \includegraphics[width=0.95\textwidth]{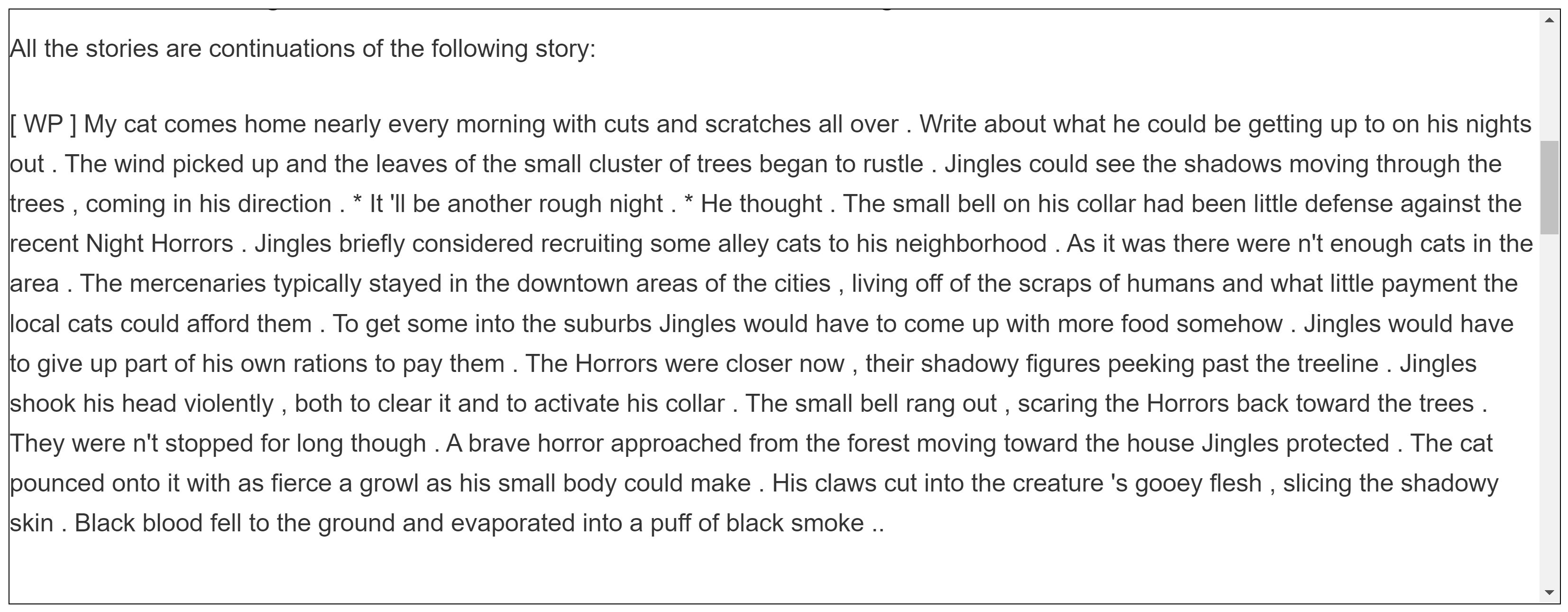}
  \caption{The story prompt show to AWS Turkers.}
  \label{fig:story_cont_story_prompt}
\end{figure*}

\begin{figure*}[htbp]
  \centering
  \includegraphics[width=0.95\textwidth]{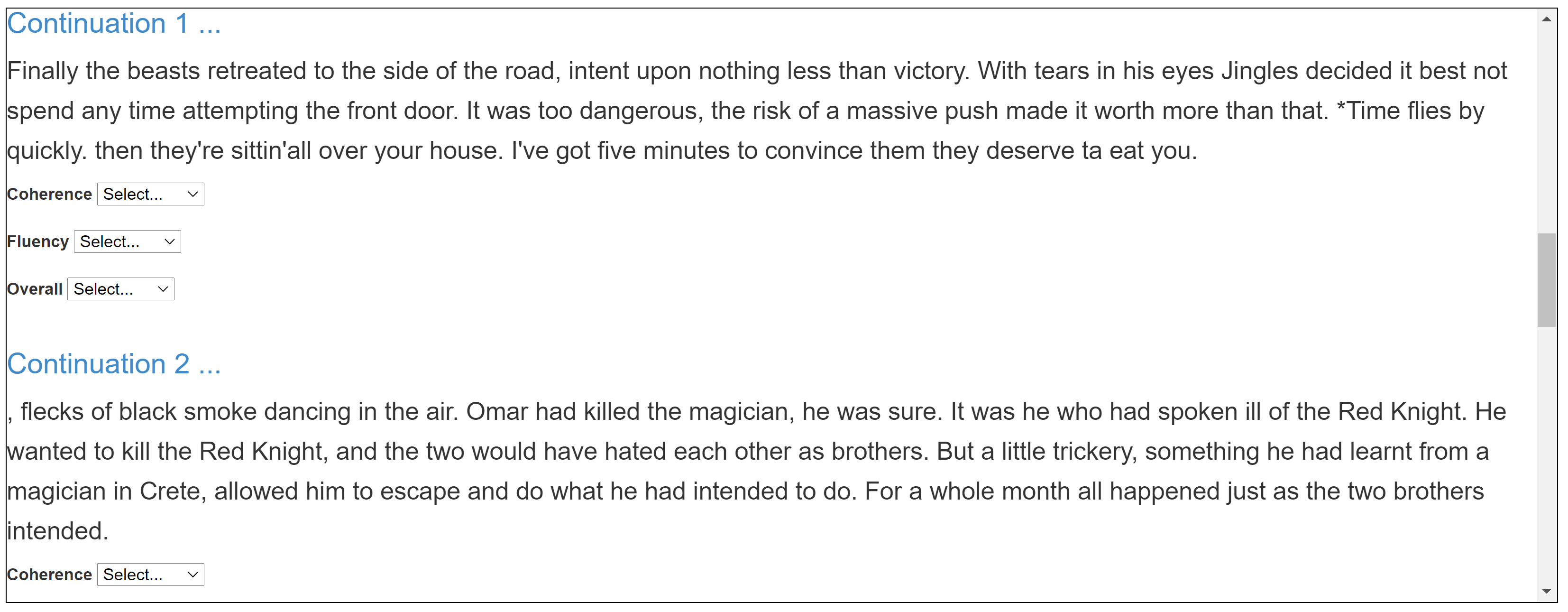}
  \caption{Showing the continuation and dropdown choices.}
  \label{fig:story_cont_select}
\end{figure*}

\begin{figure*}[htbp]
  \centering
  \includegraphics[width=0.95\textwidth]{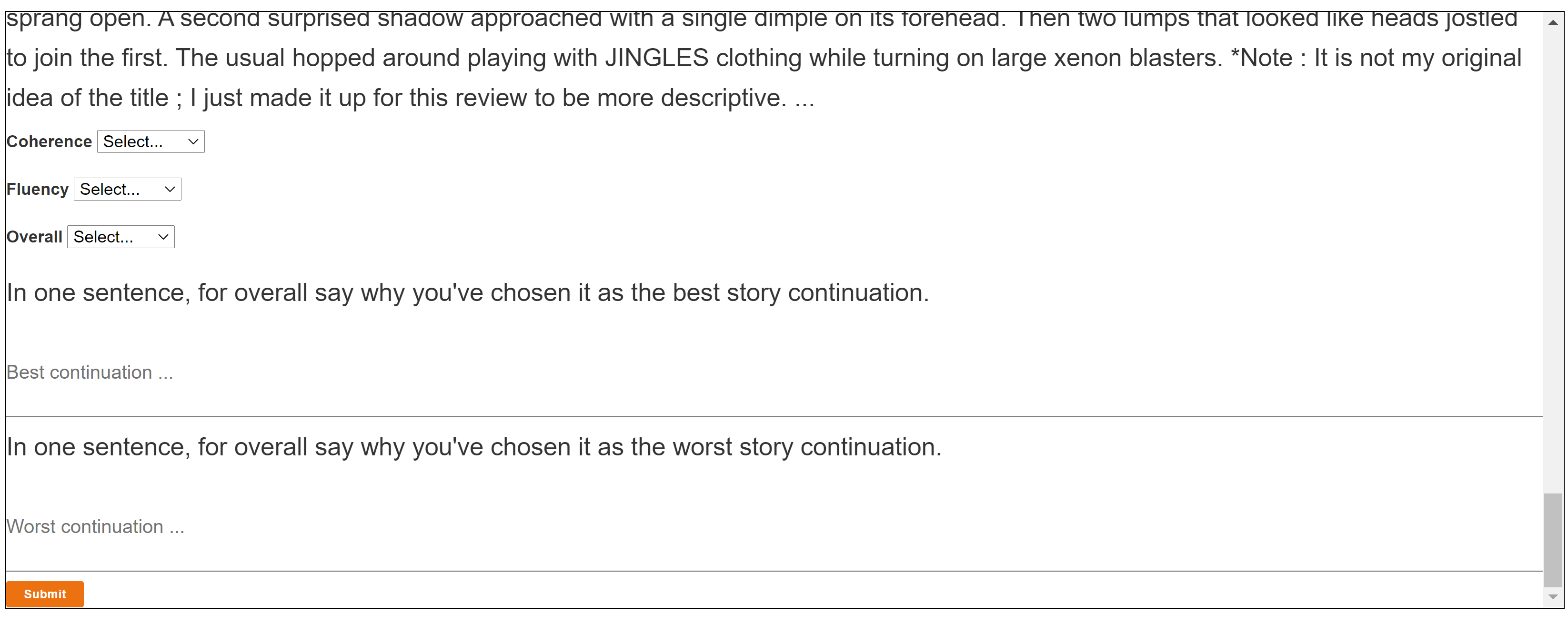}
  \caption{Checker validation question and the submit button.}
  \label{fig:story_cont_submit}
\end{figure*}

\textit{The three figures \ref{fig:story_cont_story_prompt},\ref{fig:story_cont_select}, \ref{fig:story_cont_submit} are screenshots for the AWS task showing the story prompt, a ranking box for one continuation, and the submission questions.}

\subsection{Example Generated Stories}

\label{sec:example1}

\textit{The following are two random examples of evaluated stories and the continuations generated by each model. These examples are presented in a fixed order, but in the real task the different model continuations were shuffled.}

\paragraph{Prompt} [ WP ] A woman investigates her husband 's murder ... while receiving insufferable vague help from his ghost . `` Would you like a blanket Mrs . Everson ? '' the police officer asked in a calm and collected voice . `` No , im fine thank you '' she responded , holding back her tears . The police officer nodded and made his way over to the group of officers surrounding the body on the floor . Mrs. Everson sat , quietly waiting for the next officer to ask her a question , `` leave '' a voice appeared in her ear as if emanating from the wall behind her . A shiver ran down her spine and she quickly turned to see the blank wall , as she looked back a tall man stood before her , `` Mrs. Eve Everson ? '' the man requested . `` Yes '' she responded looking up through watery eyes . The man ran through the standards , he introduced himself as the local DI , and began to present her with a series of questions . `` Did your husband have any enemies Mrs . Everson ? '' . `` No '' She responded quickly , `` yes '' came the voice from the wall , she quickly turned again to see nothing once again . `` Are you alright Mrs . Everson ? '' asked the DI `` ah , yes , sorry it was just ... nothing '' she responded nervously . `` Did you notice anything suspicious leading up to the incident ? '' The DI asked leaping right back into questioning mode , `` ah no , nothing I can thi.. ''.

\paragraph{Gold} she began before the voice cut her off `` Think Eve , THINK ! '' it screamed this time , and with the added volume the voice became recognizable . Eve leapt from the chair and fell back wards onto the ground , she scrambled away from the wall , staring at it intensely , screaming out to it `` DAVE ! DAVE ! '' . ...

\paragraph{LM} -'ve noticed at all. Must have been some kind of mistake on your part... or is this just another case of mistaken identity. '' Now that he was back in his own voice, the old wizard resumed his conversation : `` The wisest of us have fallen to thinking this over. Why have not the Wise Ones declared this thing a danger? One only knows that we are unfortunate, for it does not take much skill to hunt down the Bad One. ' ` We may at least care for our own lives, as the ogresses do,'thought they. So they agreed to go with the Bad One, and the two started on their way.

\paragraph{LSTM} - That's just normal stuff all the time these days. You don't talk too much after dark, what are you doing out here?! '' he said, heading to the door. Walking further into the old room the Man familiarly began to open her hair, dress and shoes, several tresses flying madly about her. He saw through the cloud, after all the time they had spent together the flame consumed him, but in all those years before, this woman still kept talking. She began to rub her hands together, feeling her hair dancing, her dresses floating, yet he remained so entranced. At last he ran them off as fast as thought, sure to leave no trail. ...

\paragraph{TD-VAE R} and rubbing his chin. The young woman began forcing past the police officers staring at her with wide eyed bewilderment and longing eyes, almost nauseous. She silently wandered over to where she went down the opposite corner next to the fence. This time however she stopped short of finding where Mrs. Everson had been hidden, very well probably upstairs though she might not be dead. She raised an eyebrow asking the man if the officer was lady like that, the young woman did not know, but he looked at her derisively for not actually knowing.

\paragraph{TD-VAE C} he replied angrily. Mrs. Everson could not hide her worry, trying to console herself that no one was paying any attention to her at all. ; Once upon a time, a magician walked down an avenue in our town. ; He asked a young woman what happened to her. ; At one point, he told the woman, you had knocked her head down.

\textit{The second example:}

\paragraph{Prompt} [ WP ] My cat comes home nearly every morning with cuts and scratches all over . Write about what he could be getting up to on his nights out . The wind picked up and the leaves of the small cluster of trees began to rustle . Jingles could see the shadows moving through the trees , coming in his direction . * It 'll be another rough night . * He thought . The small bell on his collar had been little defense against the recent Night Horrors . Jingles briefly considered recruiting some alley cats to his neighborhood . As it was there were n't enough cats in the area . The mercenaries typically stayed in the downtown areas of the cities , living off of the scraps of humans and what little payment the local cats could afford them . To get some into the suburbs Jingles would have to come up with more food somehow . Jingles would have to give up part of his own rations to pay them . The Horrors were closer now , their shadowy figures peeking past the treeline . Jingles shook his head violently , both to clear it and to activate his collar . The small bell rang out , scaring the Horrors back toward the trees . They were n't stopped for long though . A brave horror approached from the forest moving toward the house Jingles protected . The cat pounced onto it with as fierce a growl as his small body could make . His claws cut into the creature 's gooey flesh , slicing the shadowy skin . Black blood fell to the ground and evaporated into a puff of black smoke ..

\paragraph{Gold} More horrors came out , rushing for Jingles . Every young kitten knew about the Night Horrors , and as soon as they were old enough to fight they began training . Most kittens were lucky enough to have a sparing partner in their parent or their siblings . Jingles came from a large litter and he was no exception . He was ready for the horrors charging him .

\paragraph{LM} , flecks of black smoke dancing in the air. Omar had killed the magician, he was sure. It was he who had spoken ill of the Red Knight. He wanted to kill the Red Knight, and the two would have hated each other as brothers. But a little trickery, something he had learnt from a magician in Crete, allowed him to escape and do what he had intended to do. For a whole month all happened just as the two brothers intended.

\paragraph{LSTM} Jingle bells rang in his ears as he jumped away, brushing up against the blackened remains of his coat and fur Blurry lights blinded him once more and he cursed softly into his night ears, All because he feared The Grudge, the curse long gnawed on his bones They lured him back with stories about how *the streets were no place for bad men to roam wild and cruel man to terrorize a terrified girl. Little did he know they only made him afraid, and fed the paranoia with sad tales of candy shops filled with children and toddlers. He had grown so accustomed to life being the same as a petrified child, that eventually, the grudges abandoned him forever. He made himself miserable, now and again trying to hide in his own room and begging anyone who would listen to get off his case and tell the truth. ...

\paragraph{TD-VAE R} Disgusted with himself he began to throw open the door to the house. The household alarm woke him first when his private room window sprang open. A second surprised shadow approached with a single dimple on its forehead. Then two lumps that looked like heads jostled to join the first. The usual hopped around playing with JINGLES clothing while turning on large xenon blasters. *Note : It is not my original idea of the title ; I just made it up for this review to be more descriptive. ...

\paragraph{TD-VAE C} Finally the beasts retreated to the side of the road, intent upon nothing less than victory. With tears in his eyes Jingles decided it best not spend any time attempting the front door. It was too dangerous, the risk of a massive push made it worth more than that. *Time flies by quickly. then they're sittin'all over your house. I've got five minutes to convince them they deserve ta eat you.

\subsection{MTurk Study Details}

Human Evaluation consisted of 5 titles evaluated
by 5 judges each.  Turkers received $1$ dollar per hit. To ensure quality we found it necessary to use the Master Worker option and filter for workers with $> 98$ approval and at least $2000$ successful hits.

\subsection{Annotation Quality}

The main tests were the free form quality questions at the end asking for reasons why one story was the best or worst. Turkers who gave short answers such \textbf{bad} or \textbf{liked best} without being specific as per the instructions were rejected. $22\%$ on the initial run were rejected with replacements added on a subsequent run.

\section{Environment and Reproducibility}

\subsection{Setup}

The environment can be setup either via the \textit{requirements.txt} file with pip on the Anaconda \textit{environment.yaml} file, both in \textit{code.tar.gz}

\subsection{Predictions and Evaluations}

All of the scripts required to run cloze exercise are in the \textit{code.tar.gz} file. Scripts for running the main training and inference are in the \textit{slurm} directory and evaluation scripts within \textit{scripts}. More documentation is needed to be able to run these for a final code release.

The main inference and story generation code is in the \textit{knowledgeable-stories/predictors} package. The config is largely read through env variables in the python script. These need to be documented further for a full code release.

\subsection{Datasets}

For all training dataset the following preprocessing for all datasets is the same:

\begin{enumerate}
  \item Sentence splitting using \href{https://spacy.io/}{Spacy}.
  \item Stories are randomly shuffled according to a fixed seed.
  \item There is a 80/10/10 training/validation/test split.
\end{enumerate}

It should be noted that the only dataset used for Evaluation is WritingPrompts. For this corpus all the automatic evaluation examples are randomly sampled stories from the WritingPrompts existing testset. For automatic eveluation these stories are retsricted to those stories between $25$ and $75$ sentences excluding the prompt. 

\subsection{Training and Inference}

\begin{itemize}
  \item \textbf{Training}
    \begin{itemize}
        \item{Config}: The config files are in the \textit{\_config} within \textit{code.tar.gz}
        \item \textbf{Models}: The model files will be made available via Github when the anonymity period ends. The main TD-VAE conditioning model consists of $485M$ trainable parameters.
        \item \textbf{Policy}: All models were trained with batch size $10000$ instance per epoch in training and $1000$ for validation. The early stopping policy is to stop training after $3$ epochs without improvement. The learning rate is $0.01$ and halved every epoch training does not improve.
        \item \textbf{Time}: For the main TD-VAE conditioning model training took 82 days. Baselines depending in the variant 48-72.
        \item \textbf{Epochs}: Baseline model training ran for 7 epochs, again other models are similar.
    \end{itemize}
  \item \textbf{Generation}
     \begin{itemize}
     \item{Sampling}: Reranking models sampled $50$ sentence per time step and used a beam of $10$.
     \item{Computation}: The average rate for sentence generation on a single GPU $3$ sentences per second including reranking and TD-VAE conditionng.
     \end{itemize}
\end{itemize}

\subsection{Sentence Vector Analysis}
\label{sec:appendix_sentence_vectors}

\begin{table*}[]
\centering
\begin{tabular}{cccccccccc}
\toprule
\textbf{Model $\uparrow$} & \textbf{MR} & \textbf{CR} & \textbf{SUBJ} & \textbf{MPQA} & \textbf{SST-2} & \textbf{TREC} & \textbf{MRPC} & \textbf{SK-E}
& \textbf{SK-R} \\ \midrule
Skipthought    & 79.4        & 83.1        & 93.7          & 89.3          & 82.9           & 88.4          & 72.4          & 79.5   & 85.8         \\

InferSent    & 81.1 & 86.3 & 92.4 & 90.2 & 84.6  & 88.2 & \textbf{76.2} & \textbf{86.3} & \textbf{88.4}    \\

BERT [CLS]   & 78.7        & 84.8        & 94.2          & 88.2          & 84.1           & 91.4          & 71.1          & -        & 42.6    \\
S-BERT-NLI   & \textbf{84.9}        & \textbf{90.1}        & \textbf{94.5}          & 90.3          & \textbf{90.6}           & \textbf{97.4}          & 75.9          & 76.5        & 73.8    \\ \midrule
WP+SK        & 79.1        & 84.8        & 84.4          & \textbf{93.0}          & 83.8           & 90.4          & 68.7          & 69.7     & 31.5       \\
ALL+SK        & 77.0        & 82.8        & 82.6          & 92.6          & 78.8           & 86.4          & 64.5          & 65.2     & 36.4       \\
WP+C        & 78.5        & 82.9        & 82.5          & 92.9          & 85.0           & 88.2          & 67.1          & 73.9     & 39.1       \\
ALL+C        & 77.4        & 81.3        & 84.7          & 92.7          & 83.3           & 91.2          & 71.5          & 73.6     & 40.0       \\
WP+KB+SK        & 78.8        & 83.9        & 84.8          & 92.6          & 82.7           & 91.4          & 68.5          & 69.0     & 48.1       \\
ALL+KB+SK        & 78.9        & 83.4        & 85.6          & 92.7          & 82.3           & 89.2          & 75.7          & 68.9     & 38.8       \\
WP+KB+SK+C        & 77.2        & 82.6        & 84.7         & 92.8          & 82.7           & 87.4         & 71.0          & 72.9     & 60.0       \\
ALL+KB+SK+C         & 77.2        & 82.6        & 84.7         & 92.8          & 82.9           & 91.2         & 75.5          & 72.9     & 60.2       \\

\bottomrule
\end{tabular}
\caption{A variety of SentEval benchmarks.}
\label{tab:senteval_var}
\end{table*}

\begin{table*}[]
\centering
\begin{tabular}{ccccccccccc}
\toprule
\textbf{Model $\uparrow$} & \textbf{SL} & \textbf{WC} & \textbf{TD} & \textbf{TC} & \textbf{BSh} & \textbf{Ten} & \textbf{SuNum} & \textbf{ObNum} & \textbf{OMO} & \textbf{CInv} \\ \midrule
BoV-fasttext   & 66.6                            & \textbf{91.6}                            & 27.1                            & 68.1                            & 50.8                                & \textbf{89.1}                              & 82.1                                 & 79.8                                & 54.2                             & 54.8                               \\
SkipThought    & 68.1                            & 35.9                            & 33.5                            & \textbf{75.4}                            & 60.0                                & \textbf{89.1}                               & 80.5                                 & 77.1                                & 55.6                             & 67.7                               \\ \midrule

WP+SK         & 92.0                            & 27.1                            & 32.6                           & 66.5                            & 76.2 & 85.2  
& 83.7                                & 75.2                               & 57.6                                 & 64.5    \\

ALL+SK         & 92.7                            & 24.0                            & 31.9                           & 66.9                            & 73.3 & 86.9  
& 82.9                                & 73.5                               & 54.8                                 & 62.3    \\

WP+C          & 93.3                           & 37.7                            & \textbf{34.8}                          & 75.1                            & \textbf{80.2}                                & 83.7                               & 86.6                                & \textbf{78.7}                                & \textbf{59.6}                             & \textbf{65.7}                               \\

ALL+C          & 92.2                            & 33.6                            & 34.4                            & 75.0                            & 78.0                                & 84.3                               & \textbf{87.1}                                 & 78.4                                & 58.3                             & \textbf{65.7}                               \\

WP+KB+SK          & \textbf{93.5}                            & 26.3                           & 32.5                            & 68.5                            & 75.9                                & 87.8                               & 84.9                                 & 74.5                                & 56.7                             & 65.1                               \\

ALL+KB+SK          & 93.4                            & 26.2                            & 32.2                            & 67.2                            & 74.7                                & 87.7                               & 84.2                                 & 75.1                                & 56.6                             & 65.3                               \\ 

WP+KB+SK+C          & 92.4                            & 32.8                            & 34.1                            & 72.8                            & 78.1                                & 85.1                               & 86.6                                 & 76.5                                & 58.8                             & 65.6                               \\
ALL+KB+SK+C          & 92.4                            & 32.8                            & 32.8                            & 73.5                            & 78.1                                & 82.5                               & 86.4                                 & 76.5                                & 57.6                             & 65.6                    \\
 \bottomrule
\end{tabular}
\caption{The SentEval probing tasks.}
\label{tab:senteval_probe}
\end{table*}

One potential problem with using adhoc latent sentence representations versus the discussed planning approaches is it's not clear what the represent and whether these representations are competitive with discrete sentence embeddings models. To address this the sentence representations are evaluated with SentEval \citep{conneau-kiela-2018-senteval} and additional probing tasks \citep{conneau-etal-2018-cram}.

The other baseline models are  Skip-Thoughts \citep{NIPS2015_5950}, BERT [CLS] \citep{devlin-etal-2019-bert}, S-BERT \citep{reimers-gurevych-2019-sentence}, InferSent \citep{conneau-etal-2017-supervised}, as well as Fasttext and Bag-of-Vectors \citep{joulin2017bag}. S-BERT is state of the art or close to it for most of the tasks reported in table \ref{tab:senteval_var}, some models like \citep{Zhang2020SemanticsawareBF} which enriches BERT sentence representations with SRL information, and very large language models have done slightly better in some tasks. The tasks are representative of a diverse range of semantic tasks: MR (Movie Reviews), CR (Product Review), SUBJ (Subjectivity Status), MPQA (Opinion Polarity), SST2 (Sentiment Analyst), TREC (Question-time Classification), MRPC (Paraphrase Detection), SICK-E (Entailment), and SICK-R (Semantic Relatedness). The second table \ref{tab:senteval_probe} is the probing tasks of SentEval: SL (Sentence Length), WC (Word Content), TD (Tree Depth), TC (Top Constituents), BSh (Word Order), Ten (Verb Tense), SuNum (Subject Number), ObNum (Object Number), OMO (Odd Man Out), Coordination Inversion. The model notation is WP for just the WritingPrompts corpus, ALL for all other literary corpora, KB with additional SNLI and Atomic training, SK for using the Skip-Thoughts learning objective, and C for the Optimus sentence projection learning objective. Taken together these test the sentence representations on a wide of linguistic and semantic tasks. Generally the models perform well across the board, though not always state of the art but competitive. State of the art shouldn't be expected as mainly the models are trained on narrative text whereas SentEval in across abroad range of tasks so the fact the sentence vectors perform well is promising. The one weak area for all versions is SICK-R which is how semantically close to sentences are and WC which is picking the most informative word out of the sentence. Poor performance could be for several reasons: The sentences used in SentEval could just be quite different in word usage from narrative text. Probably more likely is that neither of the main loss functions are really tailored towards these tasks. The Skip-Thoughts loss is identifying neighbouring from non neighbouring sentences and so is more likely to represent local aspects of the story rather than S-BERT which samples negative example from other documents and passages and so is likely to be topically cohesive. The conditional memory loss is also trained to best generate the whole sentence from the context vector rather than pick out primary content or be able to judge relatedness. Overall, the various model combinations don't make much difference except in these tasks except where both the skip-thought and memory conditioning loss are present and the model performs substantially better. The main point of this analysis is secondary in trying to identify what the sentence embeddings represent and that they capture important semantic information rather than they achieve state of the art results. This would be unlikely anyway as it is a layer in a hierarchical model and not a dedicated sentence model and being evaluated out of domain for most of the SentEval tasks. The analysis satisfies that condition so it would be expected that useful for both ranking and conditioning on in downstream tasks.

\end{document}